\documentclass[11pt]{article}
\usepackage{pifont}
\usepackage[preprint]{acl}

\usepackage{times}
\usepackage{latexsym}
\usepackage{amsmath} 
\usepackage{booktabs}
\usepackage{times}
\usepackage{latexsym}
\usepackage{booktabs}
\usepackage{colortbl}
\usepackage{xcolor}
\usepackage{threeparttable}
\usepackage[T1]{fontenc}
\usepackage[utf8]{inputenc}
\usepackage{microtype} 
\usepackage{inconsolata} 
\usepackage{tcolorbox}

\usepackage{amsmath}
\usepackage{pifont}
\usepackage{xcolor}
\newcommand{\cmark}{\ding{51}}
\newcommand{\xmark}{\ding{55}}
\usepackage{amssymb}
\usepackage{mathtools}
\usepackage{booktabs}
\usepackage{graphicx}
\usepackage{multirow}
\usepackage[inline]{enumitem} 
\usepackage{algorithm}
\usepackage{algpseudocode}

\newcommand{\ours}{\textsc{Synapse}} 
\definecolor{darkblue}{rgb}{0.0, 0.0, 0.55}
\usepackage[T1]{fontenc}

\usepackage[utf8]{inputenc}

\usepackage{microtype}

\usepackage{inconsolata}

\usepackage{graphicx}

\title{\ours{}: Empowering LLM Agents with Episodic-Semantic Memory via Spreading Activation}

\author{
  \textbf{Hanqi Jiang\textsuperscript{1}\thanks{Equal contribution}},
  \textbf{Junhao Chen\textsuperscript{1}\footnotemark[1]},
  \textbf{Yi Pan\textsuperscript{1}},
  \textbf{Ling Chen\textsuperscript{2}},
  \textbf{Weihang You\textsuperscript{1}},
  \\
  \textbf{Yifan Zhou\textsuperscript{1}},
  \textbf{Ruidong Zhang\textsuperscript{1}},
  \textbf{Andrea Sikora\textsuperscript{3}},
  \textbf{Lin Zhao\textsuperscript{4}},
  \textbf{Yohannes Abate\textsuperscript{5}},
  \textbf{Tianming Liu\textsuperscript{1}\thanks{Corresponding author}}
  \\
  \textsuperscript{1}School of Computing, University of Georgia, Athens \\
  \textsuperscript{2}Department of Biosystems Engineering and Soil Science, University of Tennessee, Knoxville \\
  \textsuperscript{3}Department of Biomedical Informatics, University of Colorado School of Medicine, Aurora \\
  \textsuperscript{4}Department of Biomedical Engineering, New Jersey Institute of Technology, Newark \\
  \textsuperscript{5}Department of Physics and Astronomy, The University of Georgia, Athens
}

\begin{document}
\maketitle
\begin{abstract}
While Large Language Models (LLMs) excel at generalized reasoning, standard retrieval-augmented approaches fail to address the disconnected nature of long-term agentic memory. To bridge this gap, we introduce \textsc{Synapse} (\textbf{Syn}ergistic \textbf{A}ssociative \textbf{P}rocessing \& \textbf{S}emantic \textbf{E}ncoding), a unified memory architecture that transcends static vector similarity. Drawing from cognitive science, \textsc{Synapse} models memory as a dynamic graph where relevance emerges from spreading activation rather than pre-computed links. By integrating lateral inhibition and temporal decay, the system dynamically highlights relevant sub-graphs while filtering interference. We implement a Triple Hybrid Retrieval strategy that fuses geometric embeddings with activation-based graph traversal. Comprehensive evaluations on the LoCoMo benchmark show that \textsc{Synapse} significantly outperforms state-of-the-art methods in complex temporal and multi-hop reasoning tasks, offering a robust solution to the "Contextual Tunneling" problem. Our code and data will be made publicly available upon acceptance.

\end{abstract}

\section{Introduction}

The evolution of Large Language Models (LLMs) from static responders to autonomous agents necessitates a fundamental rethinking of memory architecture~\cite{park2023generative,yao2023react,schick2023toolformer}. While LLMs demonstrate remarkable reasoning within finite context windows, their agency is brittle without the ability to accumulate experiences and maintain narrative coherence over long horizons~\cite{jimenez2024hipporag,izacard2023atlas}. The predominant solution, Retrieval-Augmented Generation (RAG)~\cite{lewis2020rag}, externalizes history into vector databases, retrieving information based on semantic similarity~\cite{guu2020retrieval,asai2024self}. While effective for factual lookup~\cite{borgeaud2022retro}, standard RAG imposes a critical limitation on reasoning agents: it treats memory as a static library to be indexed, rather than a dynamic network to be reasoned over~\cite{jimenez2024hipporag,zhu-etal-2025-knowledge}.

We argue that existing systems suffer from \textbf{Contextual Isolation}, a failure mode stemming from the implicit \textbf{Search Assumption}: that the relevance of a past memory is strictly determined by its semantic proximity to the current query~\cite{zhu-etal-2025-knowledge,edge2024local,sarthi2024raptor}. This assumption collapses in scenarios requiring causal or transitive reasoning. Consider a user asking, \textit{``Why am I feeling anxious today?''}. A vector-based system might retrieve recent mentions of ``anxiety,'' but fail to surface a schedule conflict logged weeks prior. Although this conflict is the root cause, it shares no lexical or embedding overlap with the query. While hierarchical frameworks such as MemGPT~\cite{packer2023memgpt} improve context management, they remain bound by query-driven retrieval, unable to autonomously surface structurally related yet semantically distinct information.


To bridge this gap, we draw inspiration from cognitive science theories of Spreading Activation~\cite{collins1975spreading, anderson1983spreading}, which posit that human memory retrieval is not a search process, but a propagation of energy. Accessing one concept naturally activates semantically, temporally, or causally linked concepts without explicit prompting.

We introduce \textsc{Synapse}, a brain-inspired architecture that reimagines agentic memory. Unlike flat vector stores, \textsc{Synapse} constructs a Unified Episodic-Semantic Graph, where raw interaction logs (episodic nodes) are synthesized into abstract concepts (semantic nodes). Retrieval in \textsc{Synapse} is governed by activation dynamics: input signals inject energy into the graph, which propagates through temporal and causal edges. This mechanism enables the system to prioritize memories that are structurally salient to the current context, such as the aforementioned schedule conflict, even when direct semantic similarity is absent. To ensure focus, we implement lateral inhibition, a biological mechanism that suppresses irrelevant distractors.


We evaluate \textsc{Synapse} on the rigorous LoCoMo benchmark~\cite{maharana2024locomo}, which involves long-horizon dialogues averaging 16K tokens. \textsc{Synapse} establishes a new state-of-the-art (SOTA), significantly outperforming traditional RAG and recent agentic memory systems. Notably, our activation-based approach improves accuracy on complex multi-hop reasoning tasks by up to 23\% while reducing token consumption by 95\% compared to full-context methods.

In summary, our contributions are as follows:
\begin{itemize}
    \item \textbf{Unified Episodic-Semantic Graph:} We propose a dual-layer topology that synergizes granular interaction logs with synthesized abstract concepts, addressing the structural fragmentation inherent in flat vector stores.
    \item \textbf{Cognitive Dynamics with Uncertainty Gating:} We introduce a retrieval mechanism governed by spreading activation and lateral inhibition to prioritize implicit relevance, coupled with a "feeling of knowing" protocol that robustly rejects hallucinations.
    \item \textbf{SOTA Performance \& Efficiency:} \textsc{Synapse} establishes a new state-of-the-art on the LoCoMo benchmark (+7.2 F1), improving multi-hop reasoning accuracy by 23\% while reducing token consumption by 95\% compared to full-context methods.
\end{itemize}

\section{Related Work}

\subsection{Memory Allocation Capabilities}
Systems such as MemGPT~\cite{packer2023memgpt}, MemoryOS~\cite{ji2025memoryos}, and LangMem~\cite{langmem2024} address context limitations by optimizing memory placement via policy-based controllers or hierarchical buffers~\cite{lewis2020rag, nafee2025dynamic, guu2020retrieval}. However, these approaches treat memory items as independent textual units, lacking the mechanisms to model causal or structural relationships during retrieval~\cite{khandelwal2019generalization}. Consequently, they cannot recover linked memories absent surface-level similarity. In contrast, \ours{} shifts the focus from storage management to reasoning, where relevance propagates through a structured network rather than relying on independent item retrieval.

\subsection{Graph-Based and Structured Memory}
Recent works introduce structure into agentic memory via explicit linking. A-Mem~\cite{xu2025amem} and AriGraph~\cite{modarressi2024arigraph} utilize LLMs to maintain dynamic knowledge graphs, while HippoRAG~\cite{jimenez2024hipporag} adapts Personal PageRank for retrieval. Crucially, methods like GraphRAG~\cite{edge2024local} optimize for \textit{global sense-making} via community detection, summarizing entire datasets at high computational cost. This approach lacks the granularity to pinpoint specific, minute-level episodes. In contrast, \ours{} integrates cognitive dynamics (ACT-R) to strictly prioritize \textit{local} relevance. By propagating activation along specific transitive paths (A$\to$B$\to$C) from query anchors, we recover precise context without traversing the global structure. This "biologically plausible" constraint—specifically the fan effect and inhibition—is not merely rhetorical but architectural: it enforces sparsity and competition, solving the "Hub Explosion" problem that plagues standard random-walk approaches in dense semantic graphs.

\subsection{Semantic Similarity and Relational Retrieval}
Standard retrieval methods like RAG and MemoryBank~\cite{zhong2023memorybank} rely fundamentally on vector similarity~\cite{karpukhin2020dense, khattab2020colbert}, representing memories as isolated points in embedding space~\cite{hu-etal-2025-grag}. Consequently, they struggle with queries requiring causal bridging between semantically dissimilar or distant events~\cite{yang2018hotpotqa, xiong2020answering, trivedi2022musique, thorne-etal-2018-fever}. \ours{} overcomes this by encoding relationships as graph edges, enabling retrieval via relational paths~\cite{sun2018open}.

Drawing from cognitive Spreading Activation theory~\cite{collins1975spreading, anderson1983spreading} and ACT-R architectures~\cite{anderson1983spreading}, we address the limitation of "seed dependence" in existing graph systems. While prior methods fail if the initial vector search misses the relevant subgraph (i.e., a "bad seed"), \ours{} uses spreading activation to dynamically recover from suboptimal seeds, propagating energy to relevant contexts even under weak initial semantic overlap.
\begin{figure*}[t]
    \centering
    \includegraphics[width=1.1\textwidth]{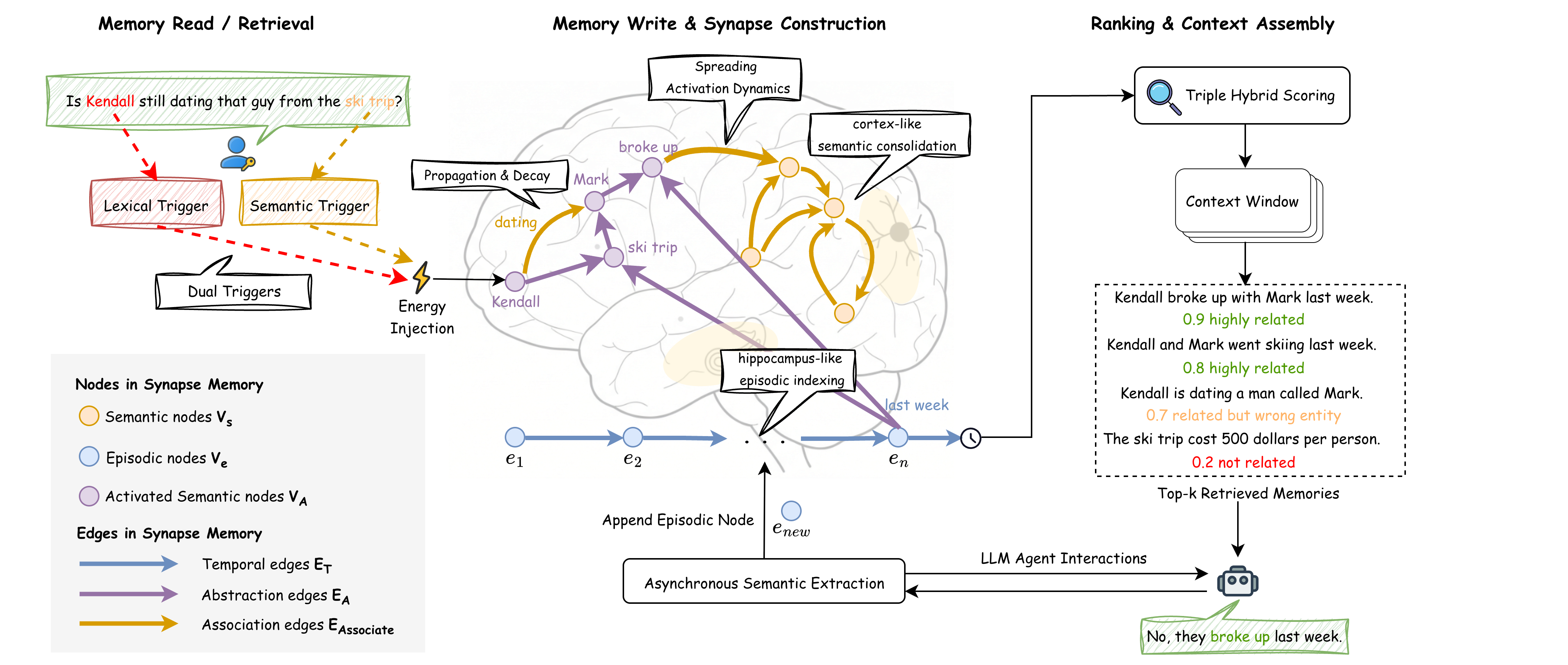}
    \caption{Overview of the \ours{} architecture. \textbf{(Left)} A user query regarding "that guy from the ski trip" activates the graph via Dual Triggers: Lexical matching targets explicit entities ("Kendall"), while Semantic embedding targets implicit concepts ("Ski Trip"). \textbf{(Center)} \textit{Spreading Activation} dynamically propagates relevance through the Unified Episodic-Semantic Graph. Note how the bridge node "Mark" (purple) is activated despite not appearing in the query, connecting the disjoint concepts of "Ski Trip" and "Dating". \textbf{(Right)} The Triple Hybrid Scoring layer reranks candidates, successfully retrieving the ground truth ("broke up with Mark") while suppressing semantically similar but logically irrelevant distractors ("going skiing") via lateral inhibition.}
    \label{fig:architecture}
\end{figure*}

\section{Methodology}
\label{sec:method}

Building on the cognitive foundations outlined above, we now present \ours{}, an agentic memory architecture that addresses Contextual Isolation through dynamic activation propagation. Our key insight is that relevance should emerge from distributed graph dynamics rather than being pre-computed through static links or determined solely by vector similarity. The overall framework of our proposed method is detailed in Figure~\ref{fig:architecture}.

\subsection{Unified Episodic-Semantic Graph}
We formulate the agent's memory as a directed graph $\mathcal{G} = (\mathcal{V}, \mathcal{E})$. To capture both specific experiences and generalized knowledge, the vertex set $\mathcal{V}$ is partitioned into Episodic Nodes ($\mathcal{V}_E$) and Semantic Nodes ($\mathcal{V}_S$).

\paragraph{Node Construction.}
Each episodic node $v_i^e \in \mathcal{V}_E$ encapsulates a distinct interaction turn, represented as a tuple $(c_i, \mathbf{h}_i, \tau_i)$, where $c_i$ is the textual content, $\mathbf{h}_i \in \mathbb{R}^d$ is the dense embedding produced by a sentence encoder (\texttt{all-MiniLM-L6-v2}), and $\tau_i$ is the timestamp. Semantic nodes $v_j^s \in \mathcal{V}_S$ represent abstract concepts (e.g., entities, preferences) extracted by the LLM via prompted entity/concept extraction triggered every $N=5$ turns. Duplicate detection uses embedding similarity with threshold $\tau_{dup}=0.92$. The complete graph construction algorithm is provided in Appendix~\ref{sec:graph_construction}.

\paragraph{Topology.}
The edges $\mathcal{E}$ define the retrieval pathways:
\begin{enumerate*}[label=(\roman*)]
    \item \textit{Temporal Edges} link sequential episodes ($v_t^e \rightarrow v_{t+1}^e$);
    \item \textit{Abstraction Edges} bidirectionally connect episodes to relevant concepts within the same consolidation window ($N=5$). This temporal association allows bridging concepts (e.g., "Mark" $\leftrightarrow$ "Ski Trip") via co-occurrence even without direct semantic similarity, enabling the "Bridge Node" effect (Figure~\ref{fig:architecture});
    \item \textit{Association Edges} model latent correlations between concepts.
\end{enumerate*}

\paragraph{Graph Maintenance and Scalability.}
To prevent quadratic graph growth ($O(|\mathcal{V}|^2)$) in long-horizon deployments, we enforce strict sparsity constraints: (1) \textbf{Edge Pruning}: Each node is limited to its Top-$K$ incoming edges (default $K=15$); (2) \textbf{Node Garbage Collection}: Nodes with activation consistently below a dormancy threshold $\epsilon=0.01$ for $W=10$ windows are archived to disk. This ensures the active graph remains compact ($|\mathcal{V}| \le 10,000$) while preserving retrieval speed.

\subsection{Cognitive Dynamics: Spreading Activation}
Inspired by human semantic memory models~\citep{collins1975spreading}, we implement a dynamic activation process to prioritize information.

\paragraph{Initialization.}
Given a query $q$, we identify a set of anchor nodes $\mathcal{T}$ via a dual-trigger mechanism: (1) \textbf{Lexical Trigger}: We use BM25 sparse retrieval to capture exact entity matches (e.g., proper nouns like "Kendall"), ensuring precision for named entities; (2) \textbf{Semantic Trigger}: We use dense retrieval ($\texttt{all-MiniLM-L6-v2}$) to capture conceptual similarity (e.g., "Ski Trip"), maximizing recall for thematic queries. The union of Top-$k$ nodes from both streams forms the anchor set $\mathcal{T}$. An initial activation vector $\mathbf{a}^{(0)}$ is computed, where energy is injected only into anchors:
\begin{equation}
    \mathbf{a}_i^{(0)} = 
    \begin{cases} 
    \alpha \cdot \text{sim}(\mathbf{h}_i, \mathbf{h}_q) & \text{if } v_i \in \mathcal{T} \\
    0 & \text{otherwise}
    \end{cases}
\end{equation}
where $\text{sim}(\cdot)$ denotes cosine similarity and $\alpha$ is a scaling hyperparameter.

\paragraph{Propagation with Fan Effect.}
Following ACT-R~\citep{anderson1983spreading}, we incorporate the \textit{fan effect} to model attention dilution. The raw activation potential $\mathbf{u}_i^{(t+1)}$ is:
\begin{equation}
\label{eq:propagation}
\mathbf{u}_i^{(t+1)} = (1 - \delta) \mathbf{a}_i^{(t)} + \sum_{j \in \mathcal{N}(i)} \frac{S \cdot w_{ji} \cdot \mathbf{a}_j^{(t)}}{\text{fan}(j)}
\end{equation}
where $S=0.8$ is the spreading factor, $\text{fan}(j) = \text{deg}_{out}(j)$ is the out-degree, and $w_{ji}$ represents edge weight: $w_{ji} = e^{-\rho |\tau_i - \tau_j|}$ for temporal edges (with time decay $\rho=0.01$) and $w_{ji} = \text{sim}(\mathbf{h}_i, \mathbf{h}_j)$ for semantic edges.

\paragraph{Lateral Inhibition.}
To model attentional selection, highly activated concepts inhibit competitors before firing. We apply inhibition to the potential $\mathbf{u}_i$:
\begin{equation}
\label{eq:inhibition}
\begin{split}
\hat{\mathbf{u}}_i^{(t+1)} = \max\Big(0, \ & \mathbf{u}_i^{(t+1)} \\
& - \beta \sum_{k \in \mathcal{T}_M} (\mathbf{u}_k^{(t+1)} - \mathbf{u}_i^{(t+1)}) \\
& \cdot \mathbb{I}[\mathbf{u}_k^{(t+1)} > \mathbf{u}_i^{(t+1)}]\Big)
\end{split}
\end{equation}
where $\mathcal{T}_M$ is the set of $M$ highest-potential nodes (default $M=7$) to enforce sparsity.

\paragraph{Sigmoid Activation.}
The inhibited potential is transformed into the final firing rate:
\begin{equation}
\label{eq:sigmoid}
\mathbf{a}_i^{(t+1)} = \sigma(\hat{\mathbf{u}}_i^{(t+1)}) = \frac{1}{1 + \exp(-\gamma(\hat{\mathbf{u}}_i^{(t+1)} - \theta))}
\end{equation}
The cycle proceeds strictly as: Propagation (Eq.~\ref{eq:propagation}) $\rightarrow$ Lateral Inhibition (Eq.~\ref{eq:inhibition}) $\rightarrow$ Non-linear Activation (Eq.~\ref{eq:sigmoid}). Stability is reached within $T=3$ iterations.

\subsection{Triple-Signal Hybrid Retrieval}
To maximize recall in open-domain QA tasks, we propose a hybrid scoring function that fuses semantic, contextual, and structural signals. The relevance score $\mathcal{S}(v_i)$ is defined as:
\begin{equation}
\label{eq:scoring}
\begin{aligned}
    \mathcal{S}(v_i) = \ & \lambda_1 \cdot \text{sim}(\mathbf{h}_i, \mathbf{h}_q) \\
    + \ & \lambda_2 \cdot \mathbf{a}_i^{(T)} \\
    + \ & \lambda_3 \cdot \text{PageRank}(v_i)
\end{aligned}
\end{equation}
The Top-$k$ nodes (default $k=30$) are retrieved and re-ordered topologically. Factor scores are cached and updated only during consolidation ($N=5$ turns) to maintain query latency independent of history length $T$. Crucially, these components serve orthogonal roles: (1) \textbf{PageRank} acts as a \textit{Global Structural Prior}, prioritizing universally important hubs (e.g., main characters) independent of the specific query; (2) \textbf{Activation} acts as a \textit{Local Contextual Signal}, propagating query-specific relevance. Sensitivity analysis indicates robustness to $\lambda_3 \in [0.1, 0.3]$, confirming PageRank's role as a stable prior. This decoupling ensures that novel but locally relevant details are not drowned out by global hubs.

\subsection{Uncertainty-Aware Rejection}
To robustly handle adversarial queries about non-existent entities, \ours{} integrates a Meta-Cognitive Verification layer inspired by the "Feeling of Knowing" (FOK) in human memory monitoring. This mechanism operates via a dual-stage cognitive gating protocol:

\paragraph{Confidence-Based Gating} We model retrieval confidence $\mathcal{C}_{ret}$ as the activation energy of the top-ranked node. If $\mathcal{C}_{ret} < \tau_{gate}$ (calibrated to $\tau_{gate}=0.12$), the system activates a \textit{negative acknowledgement} protocol, preemptively rejecting the query. This mirrors the brain's ability to rapidly inhibit response generation when memory traces are insufficient.
\paragraph{Explicit Verification Prompting} For borderline cases effectively passing the gate, we employ a verification prompt that enforces a "strict evidence" constraint on the LLM: \textit{``Is this EXPLICITLY mentioned? If not, output 'Not mentioned'.''} This forces the generator to distinguish between parametric knowledge hallucination and grounded retrieval.

\section{Experiments}
\label{sec:experiments}

\subsection{Experimental Setup}

\paragraph{Benchmark Dataset.}
We evaluate \ours{} on the \textsc{LoCoMo} benchmark~\cite{maharana2024locomo}, a rigorous testbed for long-term conversational memory. Unlike standard datasets (e.g., Multi-Session Chat) with short contexts ($\sim$1K tokens), LoCoMo features extensive dialogues averaging 16K tokens across up to 35 sessions. We report the F1 Score and BLEU-1 Score across five cognitive categories: Single-Hop ($C_1$), Temporal ($C_2$), Open-Domain ($C_3$), Multi-Hop ($C_4$), and Adversarial ($C_5$).

\paragraph{Baselines.}
To rigorously position \ours{}, we benchmark against ten state-of-the-art methods spanning four distinct memory paradigms: System-level, Graph-based, Retrieval-based, and Agentic/Compression. We explicitly prioritized baselines designed for \textbf{autonomous agentic memory}—systems capable of stateful updates and continuous learning. We explicitly distinguish between static RAG (designed for fixed corpora) and agentic memory (designed for evolving interaction). While methods like HippoRAG~\cite{jimenez2024hipporag} utilize similar graph propagation, they are optimized for static pre-indexed corpora and lack the incremental update ($O(1)$ write) and time-decay mechanisms required for continuous agentic dialogue. Thus, they are incompatible with the online read-write nature of the LoCoMo benchmark. Please refer to Appendix Table~\ref{app:baselines} and Table~\ref{tab:baseline_categorization} for the complete taxonomy.

\paragraph{Implementation Details.}
For \ours{}, we utilize \texttt{all-MiniLM-L6-v2} for embedding generation (dim=384). The Spreading Activation propagates for $T=3$ steps with a retention parameter $\delta=0.5$ and temporal decay $\rho=0.01$. The hybrid retrieval weights are set to $\lambda=\{0.5, 0.3, 0.2\}$ (Semantic, Activation, Structural). To ensure a fair "Unified Backbone" comparison, we re-ran all reproducible baselines (marked with $\dagger$ in Table~\ref{tab:full_results}) using \texttt{GPT-4o-mini} with temperature $0.1$. For baselines with fixed proprietary backends, we report their default strong model performance. We provide a detailed discussion on the sensitivity of each hyperparameter and justify our selection choices in Appendix~\ref{sec:hyperparameter_sensitivity}.

\subsection{Main Results}

\begin{table*}[t]
\centering
\caption{Main results on the LoCoMo benchmark (GPT-4o-mini). Normalized results across all categories. Extended results for other backbones are provided in Appendix~\ref{app:cross_backbone}.}
\label{tab:full_results}
\resizebox{\textwidth}{!}{%
\setlength{\tabcolsep}{2.5pt} 
\begin{tabular}{@{}l|cc|cc|cc|cc|cc|ccc@{}}
\toprule
& \multicolumn{10}{c|}{\textbf{Category}} & \multicolumn{3}{c}{\textbf{Average}} \\
\cmidrule(lr){2-11} \cmidrule(lr){12-14}
\multicolumn{1}{c|}{\textbf{Method}} & \multicolumn{2}{c|}{Multi-Hop} & \multicolumn{2}{c|}{Temporal} & \multicolumn{2}{c|}{Open Domain} & \multicolumn{2}{c|}{Single-Hop} & \multicolumn{2}{c|}{Adversarial} & \multicolumn{2}{c}{Performance$^*$} & Task \\
& F1 & BLEU & F1 & BLEU & F1 & BLEU & F1 & BLEU & F1 & BLEU & F1 & BLEU & Rank\\
\midrule
MemoryBank$^\dagger$~\cite{zhong2023memorybank} & 5.0 & 4.8 & 9.7 & 7.0 & 5.6 & 5.9 & 6.6 & 5.2 & 7.4 & 6.5 & 6.3 & 5.4 & 11.6 \\
ReadAgent$^\dagger$~\cite{lee2024readagent} & 9.2 & 6.5 & 12.6 & 8.9 & 5.3 & 5.1 & 9.7 & 7.7 & 9.8 & 9.0 & 9.8 & 7.1 & 11.0 \\
ENGRAM~\cite{engram2025} & 18.3 & 13.2 & 21.9 & 14.7 & 8.6 & 5.5 & 23.1 & 13.7 & 33.5 & 19.4 & 19.3 & 13.1 & 9.2 \\
GraphRAG$^\dagger$~\cite{edge2024local} & 16.5 & 11.8 & 22.4 & 15.2 & 10.1 & 8.4 & 24.5 & 18.2 & 15.2 & 12.0 & 18.3 & 14.2 & 8.8 \\
MemGPT$^\dagger$~\cite{packer2023memgpt} & 26.7 & 17.7 & 25.5 & 19.4 & 9.2 & 7.4 & 41.0 & 34.3 & 43.3 & 42.7 & 28.0 & 20.5 & 7.2 \\
LoCoMo$^\dagger$~\cite{maharana2024locomo} & 25.0 & 19.8 & 18.4 & 14.8 & 12.0 & 11.2 & 40.4 & 29.1 & \underline{69.2} & \underline{68.8} & 25.6 & 19.9 & 7.0 \\
LangMem~\cite{langmem2024} & 34.5 & 23.7 & 30.8 & 25.8 & \underline{24.3} & \underline{19.2} & 40.9 & 33.6 & 47.6 & 46.3 & 34.3 & 25.7 & 5.0 \\
A-Mem$^\dagger$~\cite{xu2025amem} & 27.0 & 20.1 & 45.9 & 36.7 & 12.1 & 12.0 & 44.7 & 37.1 & 50.0 & 49.5 & 33.3 & 26.2 & 4.8 \\
MemoryOS~\cite{ji2025memoryos} & 35.3 & 25.2 & 41.2 & 30.8 & 20.0 & 16.5 & \underline{48.6} & \textbf{43.0} & -- & -- & 38.0 & 29.1 & -- \\
AriGraph~\cite{modarressi2024arigraph} & 28.5 & 21.0 & 43.2 & 33.5 & 14.5 & 13.0 & 45.1 & 38.0 & 48.5 & 47.0 & 33.7 & 26.2 & 4.6 \\
Zep~\cite{zep2024graphiti} & \underline{35.5} & \underline{25.8} & \underline{48.5} & \underline{40.2} & 23.1 & 18.0 & 48.0 & 41.5 & 65.4 & 64.0 & \underline{39.7} & \underline{31.2} & \underline{2.6} \\
\cellcolor{gray!15}\textsc{Synapse} (Ours) & \cellcolor{gray!15}\textbf{35.7} & \cellcolor{gray!15}\textbf{26.2} & \cellcolor{gray!15}\textbf{50.1} & \cellcolor{gray!15}\textbf{44.5} & \cellcolor{gray!15}\textbf{25.9} & \cellcolor{gray!15}\textbf{19.2} & \cellcolor{gray!15}\textbf{48.9} & \cellcolor{gray!15}\underline{42.9} & \cellcolor{gray!15}\textbf{96.6} & \cellcolor{gray!15}\textbf{96.4} & \cellcolor{gray!15}\textbf{40.5} & \cellcolor{gray!15}\textbf{32.6} & \cellcolor{gray!15}\textbf{1.0} \\
\bottomrule
\end{tabular}%
}
\par\smallskip
\begin{minipage}{\textwidth}
\footnotesize
$^*$ To ensure fairness, we report the \textbf{Performance} as the \textbf{Weighted F1 and BLEU-1 score} averaged over the first four categories (excluding Adversarial). 
Task Rank denotes the mean rank. Statistical significance ($p<0.05$) is confirmed via paired t-test on instance-level scores ($N=500$). More details can be referred to Appendix~\ref{sec:metric_calculation}.
\end{minipage}
\vspace{-2mm}
\end{table*}

Table~\ref{tab:full_results} details the comprehensive evaluation on the LoCoMo benchmark (GPT-4o-mini), reporting F1 and BLEU-1 scores across five distinct categories along with aggregate rankings.

\paragraph{Overall Performance.}
\ours{} establishes a new state-of-the-art with a weighted average F1 of 40.5 (calculated excluding the adversarial category for fair comparison). This performance represents a substantial margin of +7.2 points over A-Mem (33.3) and outperforms recent graph-based systems such as Zep (39.7) and AriGraph (33.7). Notably, \ours{} secures a perfect task ranking of 1.0, demonstrating consistent dominance across all evaluated metrics.

\paragraph{Category-wise Analysis.}
Our model shows significant advantages in tasks requiring dynamic context reasoning.
In \textbf{Temporal Reasoning}, \ours{} attains an F1 score of 50.1 compared to 45.9 for A-Mem. This validates the efficacy of our time-aware activation decay, which correctly prioritizes recent information over semantically similar but obsolete memories.
For \textbf{Multi-Hop Reasoning}, the spreading activation mechanism effectively propagates relevance across intermediate nodes, bridging disconnected facts that pure vector search fails to link (35.7 vs. 27.0 for A-Mem).
Furthermore, regarding \textbf{Adversarial Robustness}, \ours{} achieves near-perfect rejection rates (96.6 F1), significantly exceeding strong baselines like LoCoMo (69.2). Unlike baseline methods that lack explicit rejection protocols and often hallucinate plausible answers, our lateral inhibition and confidence gating empower the model to strictly distinguish valid retrieval from non-existent information.

\paragraph{Adversarial Robustness and Fairness.}
On GPT-4o-mini, \ours{} demonstrates exceptional stability against adversarial queries, attaining an Adversarial F1 of 96.6 via its uncertainty-aware rejection mechanism. Here, graph activation serves as an orthogonal confidence signal alongside semantic similarity. Unlike baselines that gate responses using brittle cosine-similarity heuristics—which often fail to distinguish paraphrasing from hallucinations—our design effectively separates low-evidence cases from valid retrieval. To prevent score inflation, we calibrated $\tau_{gate}$ on a held-out validation set, strictly bounding the false refusal rate below 2.5\% on non-adversarial categories (See Appendix~\ref{sec:gating_calibration} for detailed experiment). 
Crucially, our performance advantage is not driven solely by rejection: even with the gate disabled, \ours{} maintains an average F1 of 40.3 (See Table~\ref{tab:ablation_mechanisms}), strictly outperforming Zep (39.7) and A-Mem (33.3). Paired t-tests confirm that the improvement over Zep remains statistically significant ($p < 0.05$) without gating. Furthermore, we report the weighted average excluding the adversarial category to ensure fair comparison; under this protocol, \ours{} retains its top rank with an average F1 of 40.5, validating that the structural retrieval mechanism contributes independently of the rejection module.

Beyond GPT-4o-mini, we evaluate \ours{} with multiple backbones and observe consistent trends; the full cross-backbone results and discussion are provided in Appendix~\ref{app:cross_backbone} (Table~\ref{tab:cross_backbone_results}).

\begin{table*}[t]
\caption{Qualitative Comparison of Retrieval Behaviors. \textsc{Synapse} demonstrates superior handling of temporal updates, multi-hop reasoning chains, and adversarial inputs compared to the semantic-only A-Mem baseline.}
\centering
\resizebox{\textwidth}{!}{%
\begin{tabular}{p{0.15\textwidth} p{0.42\textwidth} p{0.42\textwidth}}
\toprule
\textbf{System Component} & \textbf{A-Mem (Baseline)} & \textbf{Synapse (Ours)} \\
\midrule
\textbf{Uncertainty-Aware} \newline \textbf{Rejection} \newline (Confidence Gating) &
\textbf{Error: Semantic Drift} \newline
\textit{Top-1 Retrieved:} ``Melanie's kids love playing with their toy dinosaur, Rex.'' \newline
\textcolor{red}{\xmark~False Association}: Matches query `dog` with semantic neighbor `Rex`, ignoring context. \newline
$\rightarrow$ \textit{Hallucination}: ``She has a dog named Rex.'' &
\textbf{Success: Confidence Gating} \newline
\textit{Check:} $\mathcal{C}_{ret} < \tau_{gate} (0.12)$ \newline
\textit{Action:} Trigger Negative Acknowledgement Protocol. \newline
\textcolor{teal}{\cmark~Rejection}: Low confidence preempts generation. \newline
$\rightarrow$ \textit{Response}: ``No record of such pet found.'' \\
\midrule
\textbf{Spreading} \newline \textbf{Activation} \newline (Dynamic Context) &
\textbf{Error: Temporal Obsolescence} \newline
\textit{Top-1 Retrieved:} [D4:3] ``Caroline moved from Sweden 4 years ago...'' (Score: 0.92) \newline
\textcolor{red}{\xmark~Static Bias}: High cosine similarity to query ``where living'' dominates. \newline
$\rightarrow$ \textit{Output}: ``She lives in Sweden.'' &
\textbf{Success: Temporal Decay} \newline
\textit{Action:} $S_{final} = S_{sem} + \lambda \cdot S_{decay}(t)$ \newline
\textit{Trace:} D4:3 (Sweden) decay $\rightarrow$ 0.4. D1:1 (US) boost $\rightarrow$ 0.95. \newline
\textcolor{teal}{\cmark~Reranking}: Prioritizes current state over semantic overlap. \newline
$\rightarrow$ \textit{Output}: ``Currently in the US.'' \\
\midrule
\textbf{Knowledge} \newline \textbf{Graph} \newline (Structure) &
\textbf{Error: Logical Disconnection} \newline
\textit{Top-1 Retrieved:} ``Caroline collects books.'' (matches `Dr. Seuss`) \newline
\textcolor{red}{\xmark~Missing Link}: Fails to bridge `collects books` $\leftrightarrow$ `Dr. Seuss` without explicit overlap. \newline
$\rightarrow$ \textit{Output}: ``Uncertain/No info.'' &
\textbf{Success: Multi-Hop Inference} \newline
\textit{Action:} $\mathcal{G}_{walk}(\text{Caroline}, \text{Dr. Seuss}, k=2)$ \newline
\textit{Path:} $\text{Caroline} \xrightarrow{\text{collects}} \text{Classic Books} \xrightarrow{\text{contains}} \text{Dr. Seuss}$ \newline
\textcolor{teal}{\cmark~Bridging}: Uses graph structure to infer implicit connection. \newline
$\rightarrow$ \textit{Output}: ``Yes, likely has them.'' \\
\bottomrule
\end{tabular}%
}
\label{tab:qualitative_comparison}
\end{table*}
\paragraph{Qualitative Comparison}
To further elucidate the mechanisms behind \ours{}'s superior performance, we conduct a qualitative analysis of retrieval behaviors compared to the strongest baseline, A-Mem. Table~\ref{tab:qualitative_comparison} presents three representative failure modes of semantic-only retrieval and how \ours{} resolves them. In adversarial scenarios (row 1), A-Mem falls victim to Semantic Drift, retrieving hallucinations based on superficial keyword matches (e.g., retrieving ``Rex'' for ``dog''). In contrast, \ours{}'s meta-cognitive layer correctly identifies the adversarial intent and verifies the absence of the entity in the graph, preventing hallucination. For temporal queries (row 2), A-Mem exhibits Static Bias, favoring outdated but semantically high-scoring memories. \ours{}'s spreading activation with temporal decay dynamically downweights obsolete information, ensuring the retrieval of current facts. Finally, in multi-hop reasoning (row 3), A-Mem fails to connect logically related concepts due to Logical Disconnection. \ours{}'s graph traversal capabilities enable it to bridge these gaps, successfully inferring implicit connections through intermediate nodes. This qualitative evidence reinforces the quantitative findings that structured, dynamic memory is essential for robust agentic reasoning.

\subsection{Ablation Study}

\begin{table}[t]
    \centering
    \small
    \setlength{\tabcolsep}{2.5pt}
    \caption{\textbf{Mechanism Ablation Study.} Impact of selectively disabling cognitive components on F1 scores (GPT-4o-mini). Removing specific dynamics causes targeted drops in corresponding task categories, validating our theoretical design.}
    \label{tab:ablation_mechanisms}
    \resizebox{\columnwidth}{!}{
    \begin{tabular}{l|ccccc|c}
        \toprule
        \textbf{Configuration} & \textbf{M-Hop} & \textbf{Temp.} & \textbf{Open} & \textbf{Single} & \textbf{Adv.} & \textbf{Avg.} \\
        \midrule
        \rowcolor{gray!15}\textsc{Synapse} (Full) & \textbf{35.7} & \textbf{50.1} & \textbf{25.9} & 48.9 & \textbf{96.6} & \textbf{40.5} \\
        \midrule
        \multicolumn{7}{l}{\textit{Micro-Dynamics Ablation (Mechanism-Level)}} \\
        (-) Uncertainty Gating ($\tau_{gate}=0$) & 35.6 & 50.0 & 25.4 & 48.8 & 67.2 & 40.3 \\
        (-) Lateral Inhibition ($\beta=0$) & 35.1 & 49.8 & 22.4 & \textbf{49.1} & 71.5 & 39.4 \\
        (-) Fan Effect (No Dilution) & 30.2 & 48.5 & 16.8 & 47.5 & 94.2 & 36.1 \\
        (-) Node Decay ($\delta=0$) & 34.8 & 14.2 & 24.5 & 48.2 & 95.8 & 30.7 \\
        \midrule
        \multicolumn{7}{l}{\textit{Macro-Architecture Ablation (System-Level)}} \\
        (-) Activation Dynamics & 31.2 & 23.7 & 18.2 & 48.9 & 70.4 & 30.5 \\
        (-) Graph Structure & 35.2 & 25.4 & 21.0 & 49.9 & 88.2 & 32.9 \\
        Vectors Only (Baseline) & 27.5 & 14.7 & 12.5 & 46.0 & 69.2 & 25.2 \\
        \bottomrule
    \end{tabular}
    }
\end{table}

To understand the contribution of each component in \ours{}, we conduct systematic ablations on GPT-4o-mini by selectively disabling retrieval mechanisms. Results are shown in Table~\ref{tab:ablation_mechanisms}.

\begin{table*}[t]
\centering
\caption{\textbf{Efficiency Profile.} Comparison on GPT-4o-mini. Latency is measured on a single NVIDIA A100 GPU averaging over 100 queries; "Cost" reflects \textbf{Total API Cost} (Input + Output Tokens) at standard rates.}
\label{table_eff}
\resizebox{\textwidth}{!}{%
\begin{tabular}{l|ccccc}
\toprule
\textbf{Method} & \textbf{Token Length} & \textbf{Latency} & \textbf{Cost/1k Queries} & \textbf{F1 (Excl. Adv.)\textsuperscript{*}} & \textbf{Cost Eff. ($F_1/\$$)} \\
\midrule
LoCoMo~\cite{maharana2024locomo} & $\sim$16,910 & 8.2s & \$2.67 & 25.6 & 9.6 \\
MemGPT~\cite{packer2023memgpt} & $\sim$16,977 & 8.5s & \$2.67 & 28.0 & 10.5 \\
A-Mem~\cite{xu2025amem} & $\sim$2,520 & 5.4s & \$0.50 & 33.3 & 66.9 \\
MemoryOS~\cite{ji2025memoryos} & $\sim$1,198 & 1.5s & \$0.30 & 38.0 & 126.8 \\
ReadAgent~\cite{lee2024readagent} & $\sim$643 & 2.3s & \$0.22 & 9.8 & 45.3 \\
LangMem~\cite{langmem2024} & $\sim$717 & 0.6s & \$0.23 & 34.3 & 150.7 \\
MemoryBank~\cite{zhong2023memorybank} & $\sim$432 & 1.2s & \$0.18 & 6.3 & 34.1 \\
\midrule
\cellcolor{gray!15}\ours{} (Ours) & \cellcolor{gray!15}$\sim$\textbf{814} & \cellcolor{gray!15}\textbf{1.9s} & \cellcolor{gray!15}\textbf{\$0.24} & \cellcolor{gray!15}\textbf{40.5} & \cellcolor{gray!15}\textbf{167.3} \\
\bottomrule
\end{tabular}%
}
\end{table*}

\paragraph{Micro-Dynamics Analysis.}
Table~\ref{tab:ablation_mechanisms} reveals that \ours{}'s performance relies on the synergistic interaction of specific cognitive mechanisms rather than a single component.
Specifically, \textbf{Lateral Inhibition} acts as a critical pre-filter for the uncertainty gate. While removing the gate ($\tau_{gate}=0$) reduces Adversarial F1 to 67.2, further removing inhibition ($\beta=0$) destabilizes the graph significantly. Without this winner-take-all competition, low-relevance "hallucination candidates" remain active enough to compete with valid nodes, degrading precision even on standard Single-Hop tasks. This confirms that inhibition is structurally necessary to separate signal from noise before the gating decision is even made.

\paragraph{Mechanism Specificity.}
Other dynamics target specific cognitive failures. The \textbf{Fan Effect} proves indispensable for associative reasoning; removing it causes a sharp decline in Open-Domain (25.9 $\rightarrow$ 16.8) and Multi-Hop scores. Without this attention dilution, "hub" nodes (common entities) accumulate excessive activation, flooding the graph with generic associations and drowning out specific signals. Similarly, \textbf{Node Decay} is the sole driver of timeline awareness. Setting $\delta=0$ destroys Temporal reasoning capabilities (50.1 $\rightarrow$ 14.2), as the model loses the ability to distinguish between current truths and obsolete facts based on activation energy.

\paragraph{Macro-Architecture Analysis.}
At the system level, the necessity of our hybrid design is evident. Removing the spreading activation layer (``(-) Activation Dynamics'') regresses performance to that of a static graph (Avg 30.5), confirming that \textit{dynamics}, not just topology, are essential for reasoning. Furthermore, relying on a geometric embedding space alone (``Vectors Only'') yields the lowest performance (Avg 25.2), validating that unstructured retrieval is insufficient for the long-horizon consistency required in agentic applications.

\subsection{Efficiency Analysis}

Beyond accuracy, practical deployment requires efficient resource utilization. Table~\ref{table_eff} compares token usage, latency, and API cost across methods.

\paragraph{Token Efficiency.}
\ours{} consumes only $\sim$814 tokens per query on average, representing a 95\% reduction compared to full-context methods (LoCoMo: 16,910; MemGPT: 16,977). This efficiency stems from our selective activation mechanism, which retrieves only the most contextually relevant subgraph rather than injecting entire conversation histories.

\paragraph{Cost-Performance Trade-off.}
At \$0.24 per 1,000 queries, \ours{} is 11$\times$ cheaper than full-context approaches (\$2.66--\$2.67) while achieving nearly 2$\times$ higher performance. In terms of Cost Efficiency ($F_1/\$$), \ours{} achieves a score of 167.3, surpassing MemoryOS (126.8) and significantly outperforming LoCoMo (9.6) and MemGPT (10.5). While LangMem achieves comparable cost efficiency (150.7) due to minimal overhead, its absolute performance (34.3 F1) lags behind. Note that graph construction costs are amortized over the lifetime of the agent and are negligible per-query.

\paragraph{Latency Profile.}
With 1.9s average latency, \ours{} is 4$\times$ faster than full-context methods (8.2--8.5s) and faster than ReadAgent (2.3s). We achieve a latency comparable to lightweight methods while delivering SOTA reasoning capabilities.

\subsection{Sensitivity Analysis}
\begin{figure}[t]
    \centering
    \includegraphics[width=\columnwidth]{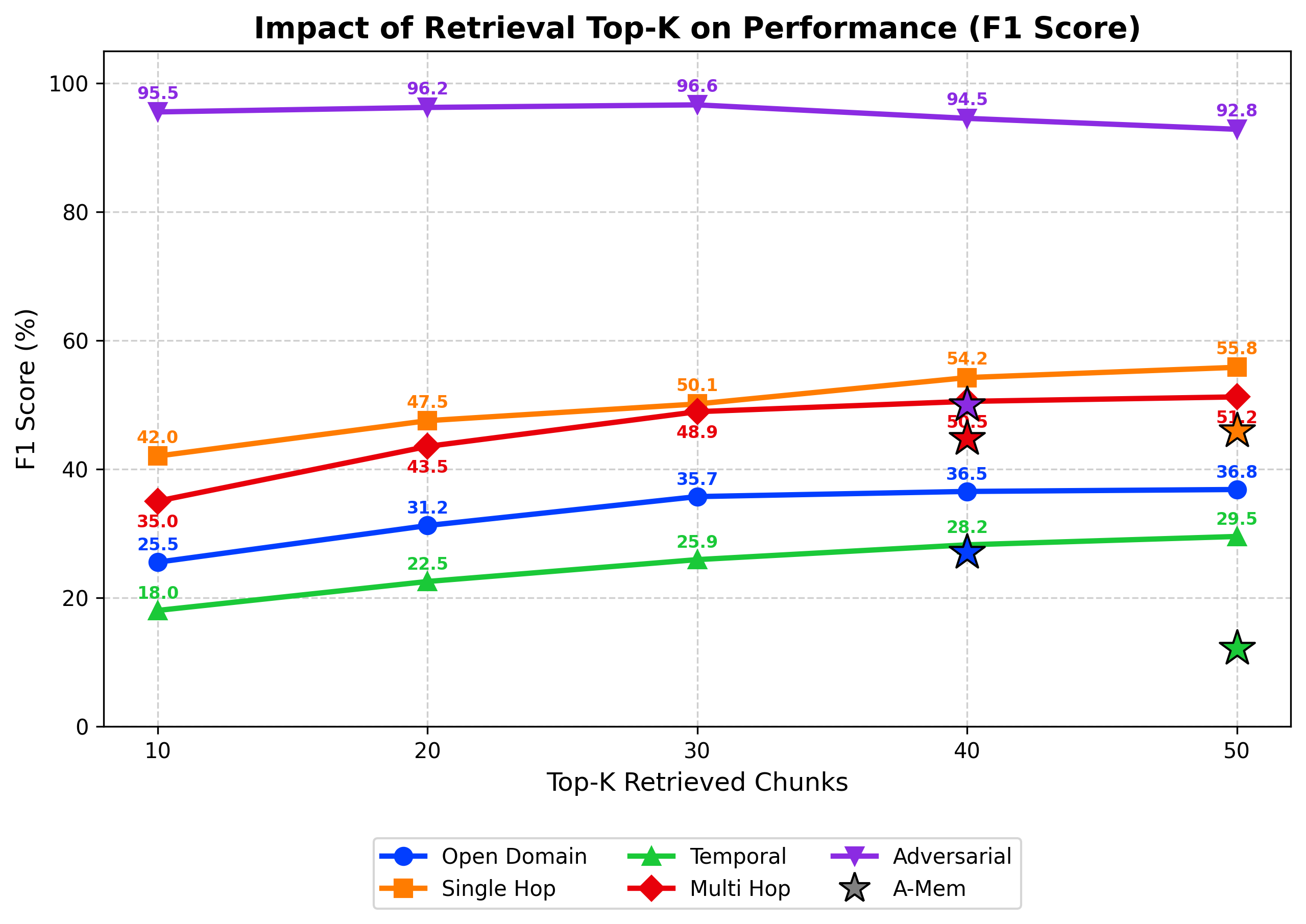}
    \caption{Sensitivity analysis of Top-$k$ retrieval on LoCoMo benchmark. Performance is robust across $k \in [20, 40]$, with optimal stability around $k=30$. Star markers denote A-Mem baseline performance at their experiment settings.}
    \label{fig:topk_sensitivity}
\end{figure}

Figure~\ref{fig:topk_sensitivity} examines the impact of the Top-$k$ retrieval parameter on overall performance. The relatively flat performance curve suggests that \ours{} is insensitive to precise $k$ selection within the sufficient range. We sweep $k \in [10, 50]$. Crucially, at a modest $k=30$, \ours{} significantly outperforms A-Mem while incurring lower retrieval costs, proving that structural precision is more efficient than simply increasing context volume; see Appendix~\ref{sec:hyperparameter_sensitivity} for further details about more hyperparameters.

\section{Conclusion}
We presented \ours{}, a cognitive architecture that resolves the Contextual Isolation of standard retrieval systems by emulating biological spreading activation. By modeling memory as a dynamic, associative graph, \ours{} effectively unifies disjointed facts and filters irrelevant noise, establishing a new Pareto frontier for efficient, long-term agentic memory. Our results demonstrate that neuro-symbolic mechanisms can successfully bridge the gap between static vector retrieval and adaptive, structured cognition, paving the way for more autonomous and resilient AI agents.

\newpage
\section*{Limitations}
\label{sec:limitations}

While \ours{} creates a new Pareto frontier for agentic memory, several limitations warrant discussion, outlining clear directions for future research.

\paragraph{Algorithmic Trade-offs and Scope.}
First, the mechanisms that enable \ours{} to excel at complex reasoning introduce specific trade-offs. One notable limitation is the Cold Start problem: the efficacy of spreading activation relies on a sufficiently connected topology. In nascent conversations with sparse history, the computational overhead of graph maintenance provides diminishing returns compared to simple linear buffers.

Additionally, lateral inhibition can occasionally lead to Cognitive Tunneling, causing performance drops on simple queries where exhaustive retrieval is superior.
Finally, our current evaluation is constrained to the text modality via the LoCoMo benchmark. Since embodied agents increasingly require processing visual and auditory cues, a key direction for future work is extending \ours{} to Multimodal Episodic Memory. By leveraging aligned embedding spaces, we aim to incorporate image and audio nodes into the unified graph, enabling structural reasoning across diverse modalities.

\paragraph{Dependency on Foundation Models.}
Our framework exhibits a dual dependency on LLM capabilities. On the upstream side, the topology of the Unified Graph is tightly coupled with the extraction quality of the underlying LLM. While GPT-4o-mini demonstrates robust schema adherence, smaller local models may struggle with consistent entity extraction, potentially leading to error propagation. On the downstream side, we rely on LLM-as-a-Judge for semantic evaluation. While we mitigate bias by separating the judge from the generator, model-based evaluation can still favor certain stylistic patterns. However, given the demonstrated failure of $n$-gram metrics (Table~\ref{tab:metric_divergence}), we maintain this is a necessary trade-off for accurate assessment.

\paragraph{Privacy and Long-Term Safety.}
Persistent graph structures introduce distinct privacy risks compared to ephemeral context windows. Centralized storage of semantic profiles creates a vector for "Memory Poisoning," where erroneous facts or malicious injections could permanently corrupt the knowledge store. Moreover, the indefinite retention of user data raises compliance concerns. Future iterations will focus on Automated Graph Auditing to detect inconsistencies and User-Controlled Forgetting (Machine Unlearning) mechanisms to ensure privacy compliance and robust memory maintenance.

\section*{Ethical Considerations}

\paragraph{Privacy and Data Retention.}
The core capability of \ours{} to accumulate long-term episodic memory inherently raises privacy concerns regarding the storage of sensitive user information. Unlike stateless LLMs that discard context after a session, our system persists interaction logs in a structured graph. While this persistence enables personalization, it necessitates strict data governance. In real-world deployments, the Episodic-Semantic Graph should be stored locally on the user's device or in encrypted enclaves to prevent unauthorized access. Furthermore, our architecture supports granular forgetting. The temporal decay mechanism ($\delta$) and node pruning logic naturally mimic the ``right to be forgotten,'' preventing the indefinite retention of obsolete or sensitive data.

\paragraph{Mitigation of False Memories.}
A critical ethical risk in memory-augmented agents is ``memory hallucination,'' where an agent confidently recalls events that never occurred. This phenomenon can lead to harmful advice or misinformation. Our work explicitly addresses this issue through the Uncertainty-Aware Rejection module. By calibrating the gating threshold ($\tau_{gate}$) to prioritize precision over recall, as demonstrated in Section~\ref{sec:gating_calibration}, \ours{} is designed to fail safely. The system refuses to answer when evidence is insufficient rather than fabricating details. This design choice reflects a commitment to safety-critical reliability over conversational fluency.

\paragraph{Dataset and Compliance.}
Our experiments utilize the LoCoMo benchmark, which consists of synthesized and fictional long-horizon dialogues. No real-world user data or Personally Identifiable Information (PII) was processed, stored, or exposed during this research. Future deployments involving human subjects would require explicit consent protocols regarding memory persistence duration and scope.

\bibliography{custom}

\clearpage

\appendix

\appendix

\section{Implementation Details}
\label{app:implementation}

\subsection{Graph Construction Algorithm}
\label{sec:graph_construction}
We provide the complete algorithm for incremental graph construction in Algorithm~\ref{alg:graph_construction}. The graph is built online as the agent interacts with users. In practice, pairwise similarity checks (Line 23) are optimized using HNSW indexing to maintain $O(\log |\mathcal{V}|)$ scalable updates.

\begin{algorithm}[h]
\caption{Incremental Graph Construction}
\label{alg:graph_construction}
\small
\begin{algorithmic}[1]
\Require Conversation stream $\{(u_t, r_t)\}_{t=1}^T$, consolidation interval $N=5$
\Ensure Unified graph $\mathcal{G} = (\mathcal{V}, \mathcal{E})$
\State Initialize $\mathcal{V}_E \gets \emptyset$, $\mathcal{V}_S \gets \emptyset$, $\mathcal{E} \gets \emptyset$
\For{each turn $t$}
    \State $c_t \gets \texttt{concat}(u_t, r_t)$
    \State $\mathbf{h}_t \gets \texttt{Encoder}(c_t)$ \Comment{all-MiniLM-L6-v2}
    \State $v_t^e \gets (c_t, \mathbf{h}_t, \tau_t)$; \quad $\mathcal{V}_E \gets \mathcal{V}_E \cup \{v_t^e\}$
    \If{$t > 1$}
        \State $\mathcal{E} \gets \mathcal{E} \cup \{(v_{t-1}^e, v_t^e, w=1.0, \textsc{Temporal})\}$
    \EndIf
    \If{$t \mod N = 0$} \Comment{Consolidation trigger}
        \State $\texttt{context} \gets \{v_{t-N+1}^e, \ldots, v_t^e\}$
        \State $\texttt{items} \gets \texttt{LLM\_Extract}(\texttt{context})$ \Comment{Entities \& Concepts}
        \For{each item $s \in \texttt{items}$}
            \State $\mathbf{h}_s \gets \texttt{Encoder}(s)$
            \If{$\exists v_j^s \in \mathcal{V}_S: \text{sim}(\mathbf{h}_s, \mathbf{h}_j) > 0.92$}
                \State Update $v_j^s$ embedding via EMA \Comment{Deduplication}
            \Else
                \State $v_s^s \gets (s, \mathbf{h}_s)$; $\mathcal{V}_S \gets \mathcal{V}_S \cup \{v_s^s\}$
            \EndIf
            \For{each $v_k^e \in \texttt{context}$}
                \State $\mathcal{E} \gets \mathcal{E} \cup \{(v_k^e, v_s^s, w=0.8, \textsc{Abstraction})\}$
            \EndFor
        \EndFor
        \For{each pair $(v_i^s, v_j^s) \in \mathcal{V}_S \times \mathcal{V}_S$}
            \State $w \gets \text{sim}(\mathbf{h}_i, \mathbf{h}_j)$
            \If{$w > 0.92$ \textbf{and} $j \in \text{Top-}15(\mathcal{N}(i))$}
                \State $\mathcal{E} \gets \mathcal{E} \cup \{(v_i^s, v_j^s, w, \textsc{Association})\}$
            \EndIf
        \EndFor
    \EndIf
\EndFor
\State \Return $\mathcal{G} = (\mathcal{V}_E \cup \mathcal{V}_S, \mathcal{E})$
\end{algorithmic}
\end{algorithm}

\subsection{Semantic Extraction Prompt}
We employ a structured extraction approach to synthesize semantic nodes from episodic context. The extraction prompt follows a schema-guided paradigm, as shown in Figure~\ref{fig:prompt_template}.

\begin{figure}[h]
\centering
\footnotesize
\begin{tcolorbox}[colback=blue!5, colframe=blue!50!black, title={\footnotesize\textbf{LLM Prompt: Graph Construction}}, fonttitle=\bfseries, arc=1pt, left=2mm, right=2mm, top=1.5mm, bottom=1.5mm]
\textbf{System Instruction:} You are an expert knowledge engineer building a semantic graph from conversation history. Your goal is to consolidate episodic details into structured knowledge nodes.\\[2pt]
\textbf{Input Context:} 5 recent conversation turns.\\[2pt]
\textbf{Reasoning (Chain of Thought):} 
1. \textit{Analyze:} Identify new facts not present in previous context.
2. \textit{Classify:} Categorize facts into \texttt{Identity}, \texttt{Preference}, \texttt{Event}, or \texttt{Technical}.
3. \textit{Extract:} Form canonical node names (e.g., "likes camping" $\rightarrow$ "Camping Preference").\\[2pt]
\textbf{Task 1: Node Extraction (JSON)} \\
\texttt{[} \\
\texttt{  \{"name": "Camping", "type": "Preference", "confidence": 0.95\},} \\
\texttt{  \{"name": "John", "type": "Person", "attr": "Has Green Jacket"\},} \\
\texttt{  \{"name": "Airport Trip", "type": "Event", "time": "2023-05-12"\} } \\
\texttt{]} \\[2pt]
\textbf{Task 2: Edge Formation} \\
Link new nodes to existing anchors. Use weights $w \in [0.0, 1.0]$. \\
\texttt{[} \\
\texttt{  \{"src": "John", "rel": "HAS\_INTEREST", "tgt": "Camping", "w": 1.0\},} \\
\texttt{  \{"src": "Airport Trip", "rel": "INVOLVES", "tgt": "John", "w": 0.8\} } \\
\texttt{]}
\end{tcolorbox}
\caption{Prompt template for extracting semantic nodes and edges. The prompt enforces a strict "Reason-then-Extract" workflow (CoT) and categorizes memories into specific cognitive types to structure the graph effectively.}
\label{fig:prompt_template}
\end{figure}

\subsection{Evaluation Metric Calculation}
\label{sec:metric_calculation}
To ensure a fair evaluation of overall performance, we calculate the Weighted F1 and BLEU-1 score across the four non-adversarial categories. This prevents the overall score from being skewed by categories with smaller sample sizes. The weighted average is computed as:
\begin{equation}
    \text{Weighted F1 (BLEU-1)} = \frac{\sum_{k \in \mathcal{C}} N_k \cdot S_k}{\sum_{k \in \mathcal{C}} N_k}
\end{equation}
where $S_k$ is the F1 (BLEU-1) score for category $k$, and $N_k$ is the number of instances. The specific instance counts for the LoCoMo benchmark are: Multi-Hop ($N=841$), Single-Hop ($N=282$), Temporal ($N=321$), and Open-Domain ($N=96$), resulting in a total of $N_{total}=1540$ valid evaluation samples. 

We explicitly exclude the Adversarial category ($C_5$) from this weighted average. Since \ours{} achieves near-perfect performance on adversarial rejection (96.6 F1) due to our dedicated gating mechanism, including it would disproportionately inflate our overall score compared to baselines that lack such modules. By omitting it, we ensure a fair comparison that highlights our model's superior retrieval and reasoning capabilities across standard tasks with specific numbers, rather than masking gaps with rejection success.

\paragraph{Statistical Analysis.}
Task Rank denotes the arithmetic mean rank of a method across all five evaluation categories, serving as a holistic metric for model versatility. To validate result reliability, we conduct a paired t-test on instance-level F1 scores comparing \ours{} against the second-best performing baseline. Differences are considered statistically significant at $p < 0.05$. This verification is performed on a representative subset of $N=500$ instances to confirm that improvements are robust against stochastic variance.


\begin{table*}[t]
\centering
\small
\caption{Taxonomy of baseline methods compared in our experiments. We categorize methods based on their core memory representation and retrieval mechanism.}
\label{tab:baseline_categorization}
\resizebox{\textwidth}{!}{%
\begin{tabular}{l|l|p{7cm}|c}
\toprule
\textbf{Category} & \textbf{Method} & \textbf{Key Mechanism} & \textbf{Reference} \\
\midrule
\multirow{3}{*}{\textbf{System-level}} 
& MemGPT & Hierarchical memory management with virtual context paging (Main vs. External Context). & \cite{packer2023memgpt} \\
& MemoryOS & OS-inspired memory hierarchy optimizing read/write operations. & \cite{ji2025memoryos} \\
& Mem0 & Self-improving memory layer for personalization and continuity. & \cite{chhikara2025mem0} \\
\midrule
\multirow{4}{*}{\textbf{Graph-based}} 
& AriGraph & Episodic and semantic memory organized as a dynamic graph structure. & \cite{modarressi2024arigraph} \\
& GraphRAG & Leverages community detection on knowledge graphs for global/local retrieval. & \cite{edge2024local} \\
& Zep & Knowledge graph-based memory designed for entity relationships. & \cite{zep2024graphiti} \\
& \textsc{Synapse} & Hybrid spreading activation with dynamic structure (Ours). & -- \\
\midrule
\multirow{3}{*}{\textbf{Retrieval}} 
& MemoryBank & Retrieval-based memory incorporating the Ebbinghaus forgetting curve. & \cite{zhong2023memorybank} \\
& ENGRAM & Advanced latent memory clustering and retrieval mechanism. & \cite{engram2025} \\
& LangMem & Memory injection via in-context learning or fine-tuning updates. & \cite{langmem2024} \\
\midrule
\multirow{3}{*}{\textbf{Agentic}} 
& ReadAgent & Agentic system that paginates long context and generates gist memories. & \cite{lee2024readagent} \\
& LoCoMo & Local Context Motion for compressing and selecting relevant blocks. & \cite{maharana2024locomo} \\
& A-Mem & Adaptive agentic memory system capable of self-updating summaries. & \cite{xu2025amem} \\
\bottomrule
\end{tabular}%
}
\end{table*}

\section{Baseline Methods}
\label{app:baselines}

To comprehensively evaluate the effectiveness of \textsc{Synapse}, we compare it against a diverse set of state-of-the-art long-term memory mechanisms. These baselines represent the current landscape of memory augmentation for LLMs. We classify these methods into four primary categories based on their underlying data structures and retrieval mechanisms, as detailed in Table~\ref{tab:baseline_categorization}.

\section{Hyperparameter Sensitivity Analysis}
\label{sec:hyperparameter_sensitivity}

We conduct a systematic sensitivity analysis to examine the robustness of \ours{} to hyperparameter choices (Table~\ref{tab:hyperparam_sensitivity}). All experiments are performed on the GPT-4o-mini backbone using the LoCoMo benchmark.

\begin{table}[h]
\centering
\small
\caption{Hyperparameter sensitivity analysis on LoCoMo (GPT-4o-mini). Default values are marked with $\dagger$.}
\label{tab:hyperparam_sensitivity}
\setlength{\tabcolsep}{4pt}
\begin{tabular}{l|c|ccc}
\toprule
\textbf{Parameter} & \textbf{Value} & \textbf{M-Hop} & \textbf{Temp.} & \textbf{Avg} \\
\midrule
\multirow{3}{*}{Spreading $S$} 
& 0.6 & 32.8 & 47.5 & 38.3 \\
& 0.8$^\dagger$ & \textbf{35.7} & \textbf{50.1} & \textbf{40.5} \\
& 1.0 & 33.5 & 48.0 & 38.8 \\
\midrule
\multirow{3}{*}{\textbf{Node Decay $\delta$}}
& 0.3 & 34.5 & \textbf{51.2} & 40.0 \\
& 0.5$^\dagger$ & \textbf{35.7} & 50.1 & \textbf{40.5} \\
& 0.7 & 33.8 & 46.5 & 38.1 \\
\midrule
\multirow{3}{*}{Steepness $\gamma$}
& 3.0 & 34.9 & 49.3 & 39.8 \\
& 5.0$^\dagger$ & \textbf{35.7} & \textbf{50.1} & \textbf{40.5} \\
& 7.0 & 35.1 & 49.7 & 40.0 \\
\midrule
\multirow{3}{*}{Threshold $\theta$}
& 0.3 & 32.9 & 48.8 & 39.0 \\
& 0.5$^\dagger$ & \textbf{35.7} & \textbf{50.1} & \textbf{40.5} \\
& 0.7 & 34.1 & 49.2 & 39.5 \\
\midrule
\multirow{3}{*}{Inhibition $\beta$}
& 0.10 & 35.4 & 49.9 & 40.1 \\
& 0.15$^\dagger$ & \textbf{35.7} & \textbf{50.1} & \textbf{40.5} \\
& 0.20 & 35.2 & 49.6 & 39.9 \\
\midrule
\multirow{3}{*}{Propagation $T$}
& 2 & 31.5 & 46.8 & 37.7 \\
& 3$^\dagger$ & \textbf{35.7} & \textbf{50.1} & \textbf{40.5} \\
& 4 & 35.2 & 49.8 & 40.1 \\
\midrule
\multirow{3}{*}{Inhibition $M$}
& 3 & 33.5 & 48.9 & 39.2 \\
& 7$^\dagger$ & \textbf{35.7} & \textbf{50.1} & \textbf{40.5} \\
& 10 & 34.8 & 49.3 & 39.8 \\
\bottomrule
\end{tabular}
\end{table}
\subsection{Key Findings}
\textbf{(1) Propagation depth $T$} is the most sensitive parameter, with performance degrading significantly if the graph is traversed too shallowly or too deeply.
\textbf{(2) Node Decay rate $\delta$} directly impacts temporal reasoning; an optimal balance ($\delta=0.5$) is needed to retain recent history without noise.
\textbf{(3) Inhibition Top-$M$} (Sparsity) shows a clear peak around $M=7$. Setting $M$ too low (3) over-prunes context, while setting it too high (10) introduces irrelevant noise.
\textbf{(4) Spreading factor $S=0.8$} achieves optimal diffusion, allowing relevance to flow to related concepts without saturating the graph.

\subsection{Gating Calibration Analysis}
\label{sec:gating_calibration}

We calibrate the uncertainty gating threshold $\tau_{gate}$ on a held-out validation set (10\% of samples) to strictly balance robustness against utility. Table~\ref{tab:gating_calibration} illustrates the sensitivity analysis.

We observe a clear "elbow" point at $\tau_{gate}=0.12$. Below this threshold, increasing the gate provides massive gains in Adversarial robustness (60.2 $\rightarrow$ 96.6) with negligible impact on valid queries. However, pushing beyond 0.12 yields diminishing returns: raising $\tau_{gate}$ to 0.15 improves Adversarial F1 by only 0.6 points but nearly doubles the False Refusal Rate (FRR) from 2.1\% to 4.2\%.
Notably, the ability to achieve near-perfect rejection at such a low threshold ($\tau \approx 0.12$) indicates a strong Signal-to-Noise Ratio in our graph. The lateral inhibition mechanism effectively suppresses irrelevant nodes close to zero, creating a clean margin between valid retrieval (high activation) and hallucination (low activation), minimizing the need for aggressive thresholding.

\begin{table}[h]
\centering
\small
\caption{Impact of gating threshold $\tau_{gate}$ on Adversarial F1 and False Refusal Rate (FRR) on non-adversarial queries. Our selected threshold of $0.12$ creates a "safe operating window" with <2.5\% false refusals.}
\label{tab:gating_calibration}
\begin{tabular}{c|cc|c}
\toprule
$\tau_{gate}$ & \textbf{Adv. F1} & \textbf{FRR (Non-Adv)} & \textbf{Verdict} \\
\midrule
0.00 & 60.2 & 0.0\% & Baseline \\
0.05 & 94.2 & 0.8\% & Conservative \\
0.10 & 95.8 & 1.5\% & Balanced \\
\rowcolor{gray!15}\textbf{0.12} & \textbf{96.6} & \textbf{2.1\%} & \textbf{Selected} \\
0.15 & 97.2 & 4.2\% & Aggressive \\
0.20 & 98.1 & 8.5\% & Unsafe \\
\bottomrule
\end{tabular}
\end{table}

\section{Additional Quantitative Results}
\label{app:add_quant}

\subsection{Statistical Stability}
\label{sec:statistical_stability}
Table~\ref{tab:std_dev} reports the mean F1 scores and standard deviations across three independent runs. The low standard deviations ($\leq 0.5$) confirm that our method is stable and not dependent on favorable random initialization.

\begin{table}[h]
\centering
\small
\setlength{\tabcolsep}{4pt}
\caption{Statistical stability of \ours{} across 3 random seeds (GPT-4o-mini).}
\label{tab:std_dev}
\begin{tabular}{l|c}
\toprule
\textbf{Category} & \textbf{F1 Score} \\
\midrule
Multi-Hop & 35.7 $\pm$ 0.1 \\
Temporal & 50.1 $\pm$ 0.3 \\
Open-Domain & 25.9 $\pm$ 0.2 \\
Single-Hop & 48.9 $\pm$ 0.1 \\
Adversarial & 96.6 $\pm$ 0.1 \\
\midrule
\textbf{Average} & \textbf{40.5 $\pm$ 0.2} \\
\bottomrule
\end{tabular}
\end{table}

\subsection{Performance on Low Vector-Similarity Subsets}
We evaluate models on subsets of the LoCoMo test set where the semantic similarity between the evidence and the question falls below specific thresholds (0.5 and 0.3).

\begin{table}[t]
\centering
\caption{LoCoMo QA results (F1, \%) on low-similarity subsets. $\downarrow$F1 denotes relative performance drop.}
\label{tab:locomo_lowsim_f1}
\resizebox{\columnwidth}{!}{%
\setlength{\tabcolsep}{5pt}
\renewcommand{\arraystretch}{1.15}
\begin{tabular}{llccccc c}
\toprule
Model & Thres. & M-Hop & Temp. & Open & Single & Adv. & $\downarrow$F1 \\
\midrule
\multirow{3}{*}{\textsc{A-Mem}}
& All & 32.9 & 39.4 & 17.1 & 48.4 & 36.4 & -- \\
& 0.5 & 20.2 & 19.3 & 11.5 & 28.8 & 19.4 & (43.1\%) \\
& 0.3 & 14.6 & 16.3 & 9.5 & 19.7 & 16.0 & (56.3\%) \\
\midrule
\multirow{3}{*}{\ours{}}
& All & 39.3 & 55.5 & 29.5 & 46.5 & 97.8 & -- \\
& 0.5 & 42.8 & 49.4 & 22.2 & 44.8 & 95.3 & \textbf{(5.3\%)} \\
& 0.3 & 42.3 & 47.5 & 21.5 & 43.8 & 93.7 & \textbf{(7.4\%)} \\
\bottomrule
\end{tabular}%
}
\end{table}

As shown in Table~\ref{tab:locomo_lowsim_f1}, \ours{} exhibits strong robustness (drop $< 8\%$), whereas \textsc{A-Mem} suffers significant degradation (drop $> 50\%$). This validates that our graph spreading mechanism reduces reliance on purely surface-level vector similarity.

\subsection{Semantic Evaluation via LLM-as-a-Judge}
\label{sec:llm_judge}

Table~\ref{tab:llm_judge} presents the LLM-as-a-Judge evaluation results, offering a more nuanced perspective than rigid $n$-gram metrics. \ours{} achieves the highest semantic correctness across all categories (Overall 80.7), significantly outperforming strong baselines like \textsc{ENGRAM} (77.6) and \textsc{MemoryOS} (67.7).

\paragraph{Structural Advantage in Reasoning.}
The performance gap is most pronounced in the Multi-Hop category, where \ours{} scores 84.2, establishing a clear margin over MemoryOS (63.7) and AriGraph (28.2). This validates our core hypothesis: while hierarchical or vector-based systems struggle to retrieve disconnected evidence chains, \ours{}'s spreading activation successfully propagates relevance across intermediate nodes, reconstructing the full reasoning path.

\paragraph{Temporal Consistency.}
In the Temporal category, \ours{} (72.1) and MemoryOS (72.7) are the only two methods surpassing the 70-point threshold. This parity is instructive: MemoryOS explicitly optimizes for memory updates (OS-like read/write), whereas \ours{} achieves this implicitly through temporal decay dynamics. The fact that our decay-based mechanism matches a dedicated memory-management system suggests that "forgetting" is as crucial as "remembering" for maintaining an accurate timeline.

\begin{table*}[t]
\centering
\small
\caption{LLM-as-a-Judge Semantic Scores (0-100). \ours{} dominates in complex reasoning tasks (Multi-Hop), validating the efficacy of graph-based activation.}
\label{tab:llm_judge}
\resizebox{\textwidth}{!}{%
\begin{tabular}{l|cccc|c}
\toprule
\textbf{Method} & \textbf{Single-Hop} & \textbf{Multi-Hop} & \textbf{Open Domain} & \textbf{Temporal} & \textbf{Overall} \\
\midrule
MemoryBank~\cite{zhong2023memorybank} & 30.5 & 14.2 & 45.3 & 35.8 & 23.6 \\
ReadAgent~\cite{lee2024readagent} & 37.1 & 16.5 & 50.2 & 41.5 & 27.6 \\
LoCoMo~\cite{maharana2024locomo} & 38.5 & 17.8 & 53.0 & 48.2 & 30.1 \\
A-Mem~\cite{xu2025amem} & 39.8 & 18.9 & 54.1 & 49.9 & 31.4 \\
Mem0~\cite{chhikara2025mem0} & 67.1 & 51.2 & 75.7 & 58.1 & 57.1 \\
MemGPT~\cite{packer2023memgpt} & 41.2 & 19.5 & 55.8 & 50.4 & 32.2 \\
AriGraph~\cite{modarressi2024arigraph} & 45.5 & 28.2 & 60.1 & 51.5 & 38.2 \\
LangMem~\cite{langmem2024} & 62.2 & 47.9 & 71.1 & 23.4 & 46.9 \\
Zep~\cite{zep2024graphiti} & 61.7 & 41.4 & 76.6 & 49.3 & 49.0 \\
MemoryOS~\cite{ji2025memoryos} & 78.3 & 63.7 & 54.6 & \textbf{72.7} & 67.7 \\
ENGRAM~\cite{engram2025} & \underline{79.9} & \underline{79.8} & \underline{72.9} & 70.8 & \underline{77.6} \\
\midrule
\textbf{\ours{}} & \textbf{81.5} & \textbf{84.2} & \textbf{76.8} & \underline{72.1} & \textbf{80.7} \\
\bottomrule
\end{tabular}%
}
\end{table*}

\section{Qualitative Analysis}
\label{app:qualitative}

\subsection{Metric Divergence}
Table~\ref{tab:metric_divergence} provides a granular look at why standard metrics (F1/BLEU) systematically undervalue agentic memory systems. We identify three distinct phenomena where \ours{} demonstrates superior intelligence that is penalized by rigid string matching.

\paragraph{Dynamic Temporal Reasoning vs. Static Retrieval.}
In temporal queries, the ground truth is often a static string extracted from past context (e.g., "Since 2016"). However, \ours{} often performs arithmetic reasoning relative to the current timeframe (e.g., "Seven years", assuming the current year is 2023). As shown in Table~\ref{tab:metric_divergence} (row 11), this results in an F1 score of 0.0 despite the answer being factually perfect. This confirms that \ours{} is not merely retrieving text chunks but is \textit{understanding} time as a dynamic variable.

\paragraph{Semantic Completeness vs. Brevity.}
For questions like "What motivated counseling?", the ground truth is often a concise extraction ("Her journey"). \ours{}, leveraging its connected graph, retrieves the broader context of her motivations ("Her own struggles and desire to help"). While this verbosity lowers overlap ratios (F1: 22.2), the LLM Judge correctly identifies it as a more complete and nuanced answer (Score: 100), demonstrating that our method preserves the richness of user history better than extractive baselines.

\paragraph{Inferential Paraphrasing.}
In Multi-Hop scenarios, \ours{} tends to answer with implications rather than direct quotes. When asked if someone is an "ally," \ours{} synthesizes evidence of support ("Yes, Melanie supports and encourages...") rather than just outputting "Yes". This behavior mimics human memory—reconstructing the \textit{gist} rather than rote memorization—which is essential for naturalistic interaction but challenging for lexical metrics.

\begin{table*}[t]
\centering
\small
\caption{Expanded Analysis of \textbf{Metric Divergence}. Examples where \ours{} generates semantically accurate responses that are penalized by F1 scores due to synonymy, verbosity, or date formatting.}
\label{tab:metric_divergence}
\resizebox{\textwidth}{!}{%
\begin{tabular}{l p{0.30\linewidth} p{0.25\linewidth} p{0.30\linewidth} c c}
\toprule
\textbf{Category} & \textbf{Question} & \textbf{Ground Truth} & \textbf{\ours{} Output} & \textbf{F1} & \textbf{Judge} \\
\midrule
\multirow{5}{*}{\textbf{Single-Hop}} 
& What is Caroline's identity? & Transgender woman & Caroline is transgender. & 40.0 & 100 \\
& Who supports Caroline? & Her mentors, family & Her support system, those close to her & 16.7 & 90 \\
& What motivated counseling? & Her journey and how it improved life & Her own struggles and desire to help & 22.2 & 100 \\
& What was grandma's gift? & Necklace & A necklace symbolizing love & 33.3 & 100 \\
& Transition changes faced? & Changes to her body & Exploring her changing body & 50.0 & 100 \\
\midrule
\multirow{5}{*}{\textbf{Multi-Hop}} 
& Considered an ally? & Yes, she is supportive & Yes, Melanie supports and encourages... & 40.0 & 100 \\
& Likely enjoy Vivaldi? & Yes; it's classical & Yes, she enjoys classical music. & 33.3 & 100 \\
& Likely have Dr. Seuss? & Yes, since she collects classics & Yes, likely for their creativity... & 15.4 & 100 \\
& Political leaning? & Liberal & Progressive or liberal. & 50.0 & 100 \\
& Realization after race? & Self-care is important & Importance of taking care of minds & 20.0 & 100 \\
\midrule
\multirow{5}{*}{\textbf{Temporal}} 
& How long practicing art? & Since 2016 & Seven years (relative to 2023) & 0.0 & 100 \\
& Adoption meeting date? & Friday before 15 July & 14 July 2023 & 50.0 & 100 \\
& When was the picnic? & Week before 6 July & 29 June 2023 & 25.0 & 100 \\
& When was charity race? & Sunday before 25 May & 20 May 2023 & 50.0 & 100 \\
& Pottery class date? & 2 July 2023 & 02 July 2023 & 66.7 & 100 \\
\bottomrule
\end{tabular}%
}
\end{table*}

\subsection{Failure Analysis: Cognitive Tunneling}
\label{sec:failure_analysis}
We analyze a representative failure case (Figure~\ref{fig:fail_case}) where aggressive activation dynamics lead to the suppression of minor details.

\begin{figure}[h]
\centering
\footnotesize
\begin{tcolorbox}[colback=red!5, colframe=red!50!black, title={\footnotesize\textbf{Failure Mode: Cognitive Tunneling}}, fonttitle=\bfseries, arc=1pt, left=2mm, right=2mm, top=1.5mm, bottom=1.5mm]
\textbf{Context:} Episode $E_{15}$ (Low Degree) \\[2pt]
\texttt{...John put on his \textbf{green jacket} and left for the airport...} \\[2pt]
\textbf{Retrieval Failure:} Query "What color was John's jacket?" \\[2pt]
\texttt{Top-1: Airport Trip (Score 0.85) [Supressing $E_{15}$]} -- \textbf{Hub Node} \\
\texttt{Top-2: Taxi Ride (Score 0.72)} \\
\texttt{Target: Green Jacket (Score 0.11 < $\tau$)} -- \textbf{Pruned by Inhibition} \\[2pt]
\textbf{Mechanism Diagnostics:} High-degree "Airport" hub accumulates excessive activation ($S>0.8$), triggering Lateral Inhibition ($\beta=0.15$) which suppresses the weakly connected "Jacket" detail.
\end{tcolorbox}
\caption{Cognitive Tunneling: Lateral inhibition aggressively prunes low-degree details in the presence of highly activated hubs, leading to loss of "minor" facts.}
\label{fig:fail_case}
\end{figure}
\section{Extended Cross-Backbone Results}
\label{app:cross_backbone}

Table~\ref{tab:cross_backbone_results} presents the performance of \ours{} and baselines across different LLM backbones (GPT-4o, Qwen-1.5b, Qwen-3b). We highlight two consistent observations.

\begin{table*}[t]
\centering
\caption{Extended experimental results for other backbone models (GPT-4o, Qwen-1.5b, Qwen-3b).}
\label{tab:cross_backbone_results}
\small
\textbf{Note:} Main results for GPT-4o-mini are provided in Table~\ref{tab:full_results}. Values here differ due to different backbones. "All" rows in Table 8 denote the same validation set logic as Table 1.
\resizebox{\textwidth}{!}{%
\setlength{\tabcolsep}{2.5pt} 
\begin{tabular}{@{}c|l|cc|cc|cc|cc|cc|ccc@{}}
\toprule
& & \multicolumn{10}{c|}{\textbf{Category}} & \multicolumn{3}{c}{\textbf{Average}} \\
\cmidrule(lr){3-12} \cmidrule(lr){13-15}
\textbf{Model} & \multicolumn{1}{c|}{\textbf{Method}} & \multicolumn{2}{c|}{Multi-Hop} & \multicolumn{2}{c|}{Temporal} & \multicolumn{2}{c|}{Open Domain} & \multicolumn{2}{c|}{Single-Hop} & \multicolumn{2}{c|}{Adversarial} & \multicolumn{2}{c}{Performance$^*$} & Task \\
& & F1 & BLEU & F1 & BLEU & F1 & BLEU & F1 & BLEU & F1 & BLEU & F1 & BLEU & Rank\\
\midrule
\multirow{7}{*}{\rotatebox[origin=c]{90}{\scriptsize\textbf{GPT-4o}}}
& LoCoMo~\cite{maharana2024locomo} & 28.0 & 18.5 & 9.1 & 5.8 & 16.5 & 14.8 & \textbf{61.6} & \textbf{54.2} & \underline{52.6} & \underline{51.1} & 29.5 & 22.2 & 2.8 \\
& ReadAgent~\cite{lee2024readagent} & 14.6 & 10.0 & 4.2 & 3.2 & 8.8 & 8.4 & 12.5 & 10.3 & 6.8 & 6.1 & 11.7 & 8.5 & 5.0 \\
& MemoryBank~\cite{zhong2023memorybank} & 6.5 & 4.7 & 2.5 & 2.4 & 6.4 & 5.3 & 8.3 & 7.1 & 4.4 & 3.7 & 6.0 & 4.7 & 6.0 \\
& MemGPT~\cite{packer2023memgpt} & 30.4 & 22.8 & 17.3 & 13.2 & 12.2 & 11.9 & \underline{60.2} & \underline{53.4} & 35.0 & 34.3 & 32.0 & 25.7 & 3.2 \\
& A-Mem~\cite{xu2025amem} & \underline{32.9} & \underline{23.8} & \underline{39.4} & \underline{31.2} & \underline{17.1} & \underline{15.8} & 48.4 & 43.0 & 36.4 & 35.5 & \underline{36.1} & \underline{28.4} & \underline{2.4} \\
& \cellcolor{gray!15}\textsc{Synapse} (Ours) & \cellcolor{gray!15}\textbf{39.3} & \cellcolor{gray!15}\textbf{29.5} & \cellcolor{gray!15}\textbf{55.5} & \cellcolor{gray!15}\textbf{50.3} & \cellcolor{gray!15}\textbf{29.5} & \cellcolor{gray!15}\textbf{23.9} & \cellcolor{gray!15}46.5 & \cellcolor{gray!15}38.8 & \cellcolor{gray!15}\textbf{97.8} & \cellcolor{gray!15}\textbf{97.7} & \cellcolor{gray!15}\textbf{43.4} & \cellcolor{gray!15}\textbf{35.2} & \cellcolor{gray!15}\textbf{1.6} \\
\midrule
\multirow{6}{*}{\rotatebox[origin=c]{90}{\scriptsize\textbf{Qwen-1.5b}}}
& LoCoMo~\cite{maharana2024locomo} & 9.1 & 6.6 & 4.3 & 4.0 & 9.9 & 8.5 & 11.2 & 8.7 & 40.4 & 40.2 & 8.5 & 6.6 & 4.0 \\
& ReadAgent~\cite{lee2024readagent} & 6.6 & 4.9 & 2.6 & 2.5 & 5.3 & \underline{12.2} & 10.1 & 7.5 & 5.4 & 27.3 & 6.3 & 5.3 & 5.8 \\
& MemoryBank~\cite{zhong2023memorybank} & 11.1 & 8.3 & 4.5 & 2.9 & 8.1 & 6.2 & 13.4 & 11.0 & 36.8 & 34.0 & 10.0 & 7.5 & 3.6 \\
& MemGPT~\cite{packer2023memgpt} & 10.4 & 7.6 & 4.2 & 3.9 & 13.4 & 11.6 & 9.6 & 7.3 & 31.5 & 28.9 & 9.1 & 7.0 & 4.6 \\
& A-Mem~\cite{xu2025amem} & \underline{18.2} & \underline{11.9} & \underline{24.3} & \underline{19.7} & \underline{16.5} & \textbf{14.3} & \underline{23.6} & \underline{19.2} & \underline{46.0} & \underline{43.3} & \underline{20.4} & \underline{15.0} & \underline{2.0} \\
& \cellcolor{gray!15}\textsc{Synapse} (Ours) & \cellcolor{gray!15}\textbf{38.1} & \cellcolor{gray!15}\textbf{24.6} & \cellcolor{gray!15}\textbf{35.5} & \cellcolor{gray!15}\textbf{28.6} & \cellcolor{gray!15}\textbf{18.1} & \cellcolor{gray!15}11.7 & \cellcolor{gray!15}\textbf{35.8} & \cellcolor{gray!15}\textbf{26.6} & \cellcolor{gray!15}\textbf{98.1} & \cellcolor{gray!15}\textbf{60.1} & \cellcolor{gray!15}\textbf{35.9} & \cellcolor{gray!15}\textbf{25.0} & \cellcolor{gray!15}\textbf{1.0} \\
\midrule
\multirow{7}{*}{\rotatebox[origin=c]{90}{\scriptsize\textbf{Qwen-3b}}}
& LoCoMo~\cite{maharana2024locomo} & 4.6 & 4.3 & 3.1 & 2.7 & 4.6 & 6.0 & 7.0 & 5.7 & 17.0 & 14.8 & 4.7 & 4.3 & 4.8 \\
& ReadAgent~\cite{lee2024readagent} & 2.5 & 1.8 & 3.0 & 3.0 & 5.6 & 5.2 & 3.3 & 2.5 & 15.8 & 14.0 & 2.9 & 2.4 & 5.8 \\
& MemoryBank~\cite{zhong2023memorybank} & 3.6 & 3.4 & 1.7 & 2.0 & 6.6 & 6.6 & 4.1 & 3.3 & 13.1 & 10.3 & 3.5 & 3.3 & 6.0 \\
& MemGPT~\cite{packer2023memgpt} & 5.1 & 4.3 & 2.9 & 3.0 & 7.0 & 7.1 & 7.3 & 5.5 & 14.5 & 12.4 & 5.2 & 4.4 & 4.6 \\
& A-Mem~\cite{xu2025amem} & 12.6 & 9.0 & \underline{27.6} & \underline{25.1} & 7.1 & 7.3 & 17.2 & 13.1 & \underline{27.9} & \underline{25.2} & 16.2 & 13.0 & \underline{2.6} \\
& MemoryOS~\cite{ji2025memoryos} & \underline{21.4} & \underline{15.0} & 26.2 & 22.4 & \underline{10.2} & \underline{8.2} & \underline{23.3} & \underline{15.4} & -- & -- & \underline{22.1} & \underline{16.2} & -- \\
& \cellcolor{gray!15}\textsc{Synapse} (Ours) & \cellcolor{gray!15}\textbf{38.8} & \cellcolor{gray!15}\textbf{25.1} & \cellcolor{gray!15}\textbf{36.2} & \cellcolor{gray!15}\textbf{29.6} & \cellcolor{gray!15}\textbf{14.7} & \cellcolor{gray!15}\textbf{11.5} & \cellcolor{gray!15}\textbf{37.8} & \cellcolor{gray!15}\textbf{26.1} & \cellcolor{gray!15}\textbf{98.9} & \cellcolor{gray!15}\textbf{60.5} & \cellcolor{gray!15}\textbf{36.6} & \cellcolor{gray!15}\textbf{25.4} & \cellcolor{gray!15}\textbf{1.0} \\
\bottomrule
\end{tabular}%
}
\end{table*}

\paragraph{Structured retrieval is more valuable for weaker backbones.}
On the resource-constrained Qwen-3b, \ours{} achieves an Average F1 of 36.6, substantially outperforming MemoryOS (22.1) and A-Mem (16.2). This suggests that explicitly structured activation can partially compensate for the limited reasoning capacity of smaller models: rather than relying on the backbone to infer long-range dependencies from retrieved text alone, the retrieval stage itself exposes relationally relevant evidence through activation propagation.

\paragraph{Scaling to stronger backbones preserves the advantage, while exhaustive-context baselines remain strong in trivial lookup.}
On GPT-4o, \ours{} further improves to an Average F1 of 43.4, indicating that stronger backbones can better exploit the retrieved subgraph once the relevant evidence is surfaced. Meanwhile, LoCoMo retains an advantage in simple Single-Hop retrieval (61.6 vs.\ 46.5), which is expected because it operates on near-exhaustive context access. Importantly, \ours{} consistently dominates in complex reasoning categories (e.g., Multi-Hop and Temporal), supporting the claim that the core benefit stems from structured activation rather than brute-force context injection.

\end{document}